# The Robustness and Super-Robustness of $L^p$ Estimation, when $p<1$

Qinghuai Gao

**Abstract:** In robust statistics, the breakdown point of an estimator is the percentage of outliers with which an estimator still generates reliable estimation.  The upper bound of breakdown point is 50%, which means it is not possible to generate reliable estimation with more than half outliers [1-2].

In this paper, it is shown that for majority of experiences, when the outliers exceed 50%, but if they are distributed randomly enough, it is still possible to generate a reliable estimation from minority good observations.  The phenomenal of that the breakdown point is larger than 50% is named as super robustness.   And, in this paper, a robust estimator is called strict robust if it generates a perfect estimation when all the good observations are perfect.

More specifically, the super robustness of the maximum likelihood estimator of the exponential power distribution, or $L^p$ estimation, where $p<1$, is investigated.  This paper starts with proving that $L^p$ ($p<1$) is a strict robust location estimator. Further, it is proved that $L^p$ has the property of strict super-robustness on translation, rotation, scaling transformation and robustness on Euclidean transform.

## 1.  Parameter Estimation Problem

A system that transforms an input $I$ to an output $O$ with a transformation $T$ is defined mathematically as below:

$$O = T(I) \qquad (1)$$

Estimating the transformation $T$ based on a group of input and output pairs of a system is a central and challenging problem in many pattern matching and computer vision systems. Typical examples are medical image registration, fingerprint matching, and camera model estimation.

The following concepts will be used in this paper:

**Estimator** An estimation approach to generate the system parameters based on groups of observations.

**Experiment** A group of observations that is used to generate an estimated transformation.

**Robustness** The characteristics of an estimator that the estimated transformation whose error to the ideal (best) estimation is bounded even when all the noise observations move to infinite.

**Strict robustness** The capability of an estimator that gives the perfect estimation even when the good observations are perfect but the noise observations have any possible distribution.

**Super robustness** The characteristics that an estimator generates an estimated transformation whose error to the perfect estimation is bounded even when the noise observations are majority, and they move to infinite.

**Strict super robustness** The characteristics that an estimator generates the perfect estimation when the good observations are perfect and the noise observations are majority.

**Breakdown point** The percentage of noise that a robust estimator tolerates is called its breakdown point. By "tolerates", it means no matter how the noise observations are distributed, the estimated result still has bounded error with the ideal estimated value. For robust estimators, the upper bound of breakdown point is 50%.

Even though, it is not possible that an estimator tolerates majority noise observation in any distribution, however, it is possible that an estimator tolerates the majority noise observations in special distributions, and even in the majority of the distributions.

This paper will investigate the robustness and super-robustness characteristics of $L^p$ where *p<1* on translation, Euclidean transformation and scaling transformation.

Suppose $I_1, I_2 \cdots, I_N$ are $N$ inputs of a system defined in the formulae (1), where $I_i$ is a point in an Euclidean space, and $O_1, O_2 \cdots, O_N$ are the corresponding outputs, where $O_i$ is a point in an Euclidean space that may have different dimension than the input space. For a transformation $T$, we define the difference of $O_i$ and $T(I_i)$ as

$$d(O_i, T(I_i)) \tag{2}$$

Thus, the overall difference between the observed outputs and the estimated outputs based on $T$ is

$$\sum_{i=1}^{N} d(O_i, T(I_i)) \tag{3}$$

The problem to estimate $T$ becomes that find a $T_b$, which satisfies:

$$D_{T_b} = \min_{T} \sum_{i=1}^{N} d(O_i, T(I_i)) \tag{4}$$

The minimum takes on any possible transformation $T$ in a predefined transformation group. The transformation groups that will be discussed in this paper are translation, scaling transformation and Euclidean transformation. When $d$ is Euclidean distance, it is the least square estimation.

In this paper, we use $L^p$ *(p<1)* to define the difference, that is, the difference of $O_i$ and $T(I_i)$ is

$$|O_i - T(I_i)|^p \tag{5}$$

Thus, the estimation problem is converted to that find a $T_b$ that satisfies:

$$\sum_{i=1}^{N} |O_i - T_b(I_i)|^p = \min_{T} \sum_{i=1}^{N} |O_i - T(I_i)|^p \tag{6}$$

To observe the robustness characteristics of $L^p$, we divide the observations into two groups: all observations in the first group are perfect observations: for $i = 1, 2, \cdots, n$, $O_i = T_b(I_i)$, where $T_b$ is the ideal transformation; all observations in the second group are noisy, that is, $O_i \neq T_b(I_i)$, where $i = n+1, n+2, \cdots, N$.

We denote that the number of noise observations as $M$, with $M = N - n$.

We will estimate the relation of $n$ and $M$ to understand the robustness of $L^p$ (*p<1*). In other words, how much percent of the ideal output out of the total observations still results an ideal estimation $T_b$ or a reliable estimation of $T$ when $L^p$ estimator is used.

To pursue strict robustness, we expect that $n$ satisfies

$$D_T \geq D_{T_b} \tag{7}$$

for any possible transformation $T$ in a transformation group. Or

$$\sum_{i=1}^{n}|O_i - T(I_i)|^p + \sum_{i=1}^{M}|O_{i+n} - T(I_{i+n})|^p \geq \sum_{i=1}^{M}|O_{i+n} - T_b(I_{i+n})|^p \tag{8}$$

since $O_i = T_b(I_i)$ for $i = 1, 2, \cdots\cdots, n$.

## 2. **Strict Robustness on Translation**

For arbitrary form of $T$, it is difficult or impossible to investigate the robustness of an estimator. To simplify the problem, we start with a simple transformation group: translation. For a translation, we define it as:

$$T(I) = I + a_T \tag{9}$$

For the ideal transformation $T_b$, we define it as:

$$T_b(I) = I + a_{T_b} \tag{10}$$

For the *i*-th observation in the first group, the difference between the observation and the output of a system with a transform $T$ is:

$$O_i - T(I_i) = T_b(I_i) - T(I_i) = a_{T_b} - a_T \tag{11}$$

By denoting that $d_T = |a_{T_b} - a_T|$, we have:

$$nd_T^p + \sum_{i=1}^{M}|O_{i+n} - T(I_{i+n})|^p \geq \sum_{i=1}^{M}|O_{i+n} - T_b(I_{i+n})|^p \tag{12}$$

Denote that $d_i = |O_{i+n} - T_b(I_{i+n})|$. The right side of the above inequality is

$$\sum_{i=1}^{M} d_i^p \tag{13}$$

The left side is no less than:

$$nd_T^p + \sum_{i=1}^{M}|d_i - d_T|^p \tag{14}$$

Thus, if

$$nd_T^p + \sum_{i=1}^{M}|d_i - d_T|^p \geq \sum_{i=1}^{M} d_i^p \tag{15}$$

or

$$nd_T^p \geq \sum_{i=1}^{M} d_i^p - \sum_{i=1}^{M}|d_i - d_T|^p \tag{16}$$

then $D_T \geq D_{T_b}$.

When $d_T \leq d_i$, we have

$$d_i^p - |d_i - d_T|^p \leq d_T^p \tag{17}$$

This is because $(a+b)^p < a^p + b^p$ when $p < 1$, $a > 0$ and $b > 0$.

Similarly, when $d_T > d_i$,

$$d_T > d_i, d_i^p - |d_T - d_i|^p < d_T^p \tag{18}$$

Thus, when $n \geq M$, then $D_T \geq D_{T_b}$ for any translation. That is $L^p (p < 1)$ is a strict robust location estimator.

## 3. **Strict Super Robustness on Translation**

To make estimation simpler, without loss generality, we assume that:

$$d_1 < d_2 < \cdots < d_M \tag{19}$$

We will discuss the case that $d_i = d_{i-1}$ for some of $i$-s later.

Let $s = \min_i (d_i - d_{i-1})$ and $S = \max_i (d_i - d_{i-1})$. When $d_T \in [d_k, d_{k+1})$, we divide the items into two groups:

$$\sum_{i=1}^{k}\left[d_i^p - (d_T - d_i)^p\right] + \sum_{i=k+1}^{M}\left[d_i^p - (d_i - d_T)^p\right] \tag{20}$$

For the first group where $d_i$ is no larger than $d_T$ (named TFG in this paper), we have:

$$\sum_{i=1}^{k}\left[d_i^p - (d_T - d_i)^p\right]$$

$$\leq \sum_{i=1}^{k}\left[(Si)^p - ((k-i)s)^p\right]$$

$$= (S^p - s^p)\sum_{i=1}^{k-1} i^p + S^p k^p$$

$$\leq (S^p - s^p)\int_0^k x^p\,dx + S^p k^p$$

$$= (S^p - s^p)k^{p+1}/(p+1) + S^p k^p$$

$$\leq (S^p - s^p)d_T^{p+1}/s^{p+1}(p+1) + S^p k^p \tag{21}$$

For the second group (named TSG in this paper) where $d_i > d_T$, we have:

$$\sum_{i=k+1}^{M}\left[d_i^p - (d_i - d_T)^p\right]$$

$$\leq \sum_{i=k+1}^{M}\left[d_i^p - (d_i - d_T)^p\right](d_i - d_{i-1})/s$$

$$\leq \sum_{i=k+2}^{M}\int_{d_{i-1}}^{d_i}\left[x^p - (x - d_T)^p\right]dx\Big/s + d_{k+1}^p - (d_{k+1} - d_T)^p$$

$$= \int_{d_{k+1}}^{d_M}\left[x^p - (x - d_T)^p\right]dx\Big/s + d_{k+1}^p - (d_{k+1} - d_T)^p$$

$$\leq \left(d_M^{p+1} - (d_M - d_T)^{p+1} - d_{k+1}^{p+1} + (d_k - d_T)^{p+1}\right)\Big/(s(p+1)) + d_T^p$$

$$\leq \left(d_M^{p+1} - d_T^{p+1} - \sum_{i=0}^{\infty}\binom{p+1}{i}d_M^{p+1}(-d_T/d_M)^i\right)\Big/s(p+1) + d_T^p \tag{22}$$

The formula above is valid because formulae (17), and

$$-d_{k+1}^{p+1} + (d_{k+1} - d_T)^{p+1} \leq -d_T^{p+1} \tag{23}$$

The later is deducted by $(a+b)^{p+1} > a^{p+1} + b^{p+1}$, and let $a = d_{k+1} - d_T$, and $b = d_T$.

For the last line in the formulae (22), from $i \geq 2$ on, $(-d_T/d_M)^i$ is negative for odd items and positive for even items, and $\binom{p+1}{i}$ is vice versa, so $\binom{p+1}{i}d_M^{p+1}(-d_T/d_M)^i$ is negative. Thus,

$$-\sum_{i=0}^{\infty}\binom{p+1}{i}d_M^{p+1}(-d_T/d_M)^i$$

$$= -d_M^{p+1} + (p+1)d_M^p d_T + many \quad negative \quad items \tag{24}$$

We have that

$$\sum_{i=k+1}^{M}\left[d_i^p - (d_i - d_T)^p\right] \leq \left((p+1)d_M^p d_T - d_T^{p+1}\right)/s(p+1) + d_T^p \tag{25}$$

Thus, all $n$ that satisfies

$$\begin{aligned}n &\geq (S^p - s^p)d_T/s^{p+1}(p+1) + (S/s)^p + \left((p+1)d_M^p d_T - d_T^{p+1}\right)/\left(d_T^p s(p+1)\right) + 1 \\ &= R^p + \left(d_M^p d_T^{1-p} + (R^p - 2)d_T/(p+1)\right)/s + 1\end{aligned} \tag{26}$$

guarantees that $D_T \geq D_{T_b}$, where $R = S/s$. It reaches maximum

$$n \geq p\left(\frac{1-p^2}{2-R^p}\right)^{(1-p)/p} d_M + R^p + 1 \tag{27}$$

when

$$d_T = \left(\frac{1-p^2}{2-R^p}\right)^{1/p} d_M \tag{28}$$

When $R^p \geq 2$, the right side reaches the maximum when $d_T = d_M$. The right side is no less than $M$. In this case, we can't observe anything better than strict robustness discussed above.

### 3.1 Case $R = 1$

An interesting case is that $S = s$, for which we have

$$n > p(1-p^2)^{(1-p)/p} M + 2 \tag{29}$$

Here, the condition $S = s$ means the distance from the $i$-th observed noisy output and its ideal result is $si$. No constrains on the location of the noisy output. For example, all the ideal outputs are located at the origin of the Euclidean space and the noisy outputs are located at $(si, 0, 0, \cdots, 0)$ of the first axis, which is a very biased noisy case.

Since $p(1-p^2)^{(1-p)/p} \leq 1$, we have a simple lower bound of $n$

$$n > pM + 2 \tag{30}$$

When $M$ is big, we only need about $pM$ good observations to generate the perfect estimation. The table below numerically shows the relation between the lower bound of $n$ and $M$ when $S = s$.

| $p$ | $\min(n)/M$ |
|---|---|
| 0.1 | 0.09 |
| 0.2 | 0.17 |
| 0.3 | 0.24 |
| 0.4 | 0.31 |
| 0.5 | 0.38 |
| 0.6 | 0.45 |
| 0.7 | 0.52 |
| 0.8 | 0.62 |
| 0.9 | 0.75 |
| 1 | 1.00 |

**Table 1**. $M$ and $n$ relation when $S = s$

This is a very encouraging result. For $p = 0.1$, we only need 1/11 good observations to estimated the location no matter how the noisy output distributed.

## 3.2 Case $R^p < 2$ and approaches to lower the bound of $n$

When $R^p < 2$, the maximum is reached when

$$\left(d_M^p d_T^{1-p} + (R^p - 2)d_T/(p+1)\right)' = 0 \tag{31}$$

That is, when

$$d_T = \left[(1-p^2)/(2-R^p)\right]^{1/p} d_M \tag{32}$$

the right side of inequality (27) reaches its maximum

$$p\left[(1-p^2)/(2-R^p)\right]^{(1-p)/p} d_M/s + 1 + R^p \tag{33}$$

When

$$n > p\left[(1-p^2)/(2-R^p)\right]^{(1-p)/p} d_M/s + 1 + R^p \tag{34}$$

, we have $D_T \geq D_{T_b}$ for any translation. When $R \leq (1+p^2)^{1/p}$, we have a lower bound of $n$:

$$n > p d_M/s + 1 + R^p \tag{35}$$

For some distribution of $d_i$, $n$ is smaller than $M$ so that $L^p (p < 1)$ is strict robust estimator.

We supposed that $s > 0$ in the discussion in this section. It is easy to observe that when $s = 0$ or $s$ is very small, the inequality above will be meaningless since the right side goes to infinite. We separate the noisy outputs into two sets:

$$S_1 = \{O_i \mid d_i < s\} \tag{36}$$

And let $S_2$ be the complimentary set of $S_1$.

We denote that $m = |S_1|$. For those $O_i$ in $S_1$, their contribution to the right side should be no more than $m$ (see the discussion in the section above). Now we have:

$$n > m + p\left[(1-p^2)/(2-R^p)\right]^{(1-p)/p} d_{M-m}/s + 1 + R^p \tag{37}$$

, where $d_{M-m}$ is the largest distance for $S_2$.

This inequality has two means: If we expect to observe super robustness, the group of the ideal observations must be the largest group which generates a consistent estimation. In other words, $L^p$ generates the estimation from the largest group of the observations. If a group of noisy observation consistently drags to a estimation and its size is larger than the good observation, we couldn't obtain the ideal observation, which is mathematically (and politically) reasonable.

In the case that only one input satisfies that $|d_{i+1} - d_i| = S$ and all the others have $|d_{i+1} - d_i| = s$, the estimated lower bound of $n$ is too high. In this case, let $S' = S - \lfloor S/s \rfloor + s$, and add $\lfloor S/s \rfloor - 1$ noisy observations between $O_i$ and $O_{i+1}$ whose distance is $s$. So the above estimation is lowered to:

$$n > p\left[(1-p^2)/(2-R'^p)\right]^{(1-p)/p} d_M/s + \lfloor S/s \rfloor + 1 + R'^p \tag{38}$$

, where $R' = S'/s$, and the item $\lfloor S/s \rfloor$ is used to compensate error introduced by the new observations. With this trick, we lower $R'$ to the range of [1,2). When a small percent of the observations goes to infinite, we separate them into another set so that we control $d_M$ to avoid it goes to infinite. A good estimation of $d_M$ is:

, where $\bar{s} = \sum_{i=1}^{M} s_i \Big/ M$. If the size of set that $d_i$ is very small is $m_1$, and the set for the extremely noisy observations is $m_2$, and the number of new observations we add to lower the ratio $R$ is $m_3$, we have

$$n > m_1 + m_2 + m_3 + p\left[(1-p^2)/(2-R'^p)\right]^{(1-p)/p} \bar{s}(M-m_1-m_2)/s + 1 + 2^p \quad (39)$$

Or

$$n > m_1 + m_2 + m_3 + p\bar{s}(M-m_1-m_2)/s + 1 + 2^p \quad (40)$$

when $R' \leq (1+p^2)^{1/p}$.

One special case in which $n$ is close to $M$ is when $O_i$ form a number of groups, for each group a zero error translation is defined: $O_i = T + I_i$. For this case, $d_i = 0$ in each group, which means $m_1$ can be almost as large as $M$. However, we regroup the observations by picking one from each of the group. This new grouping guarantees that in each of the group we newly create, $d_i > 0$. For each group, we have a lower bound of $n$ that is a function of $R$, $s$, and $d_M$ for the group. The lower bound of $n$ is the summation of these lower bounds, and the size of those very small groups:

$$n > m_1 + m_2 + \sum_{i=1}^{N_g} \left\{ m_i + p\left[(1-p^2)/(2-R_i'^p)\right]^{(1-p)/p} \bar{s}_i M_i/s_i + 1 + 2^p \right\} \quad (41)$$

, where $N_g$ is the number of groups after regrouping. With this analytic technique, we lower the lower bound of $n$ when the group sizes are relatively large.

Until now, we see that the most difficult case to observe super robustness is that the number of the zero-error groups, (in each of which $d_i = 0$), is small. In this case, the size of the ideal observation group has to be larger than the summation of the sizes of the other groups to guarantee to obtain the ideal transformation $T_b$. This retrogrades to the case of strict robustness we discussed in Section 2.

### 3.3 $d_i$ forms a half normal distribution

The condition $S = s$ is very restricted. Now we discuss another more general case: $\{d_1, d_2, \ldots, d_M\}$ forms a normal distribution with a mean of $\mu$ and a standard derivation of $\sigma$. Further we suppose that $\mu = 4\sigma$. We cut those noisy observations that satisfy $d_i < \sigma$ and $d_i > 7\sigma$ out. The total of those cut out is less than 1%. For the noisy observations left, we have $s = \sigma$ and $S = 7\sigma$, which means $R = 7$. It is also reasonable suppose that $d_M = M\mu$. The table below is a numerical result showing the super robustness of this case. Since the distribution of the errors varies in a large range, we have to choose relatively small $p$-s ($p$ in the range of 0.001 to 0.02) to observe the super robustness. (From discusses above, one necessary condition to observe super robustness is that $2 > R^p$. The largest $p$ that satisfies this condition when $R$ is 7 is about 0.356, which means $p$ has to be smaller than 0.356. However, this is not sufficient: the maximum $p$ which guarantee super robustness using the formula above is about 1/10 of 0.356 as shown in the table below.)

| $p$ | $\min(n)/M$ |
| --- | --- |
| 0.001 | 0.028 |
| 0.002 | 0.056 |
| 0.004 | 0.113 |
| 0.006 | 0.169 |
| 0.008 | 0.226 |
| 0.01 | 0.283 |
| 0.012 | 0.34 |
| 0.014 | 0.40 |
| 0.016 | 0.45 |
| 0.018 | 0.51 |
| 0.02 | 0.57 |
| 0.022 | 0.63 |
| 0.024 | 0.69 |
| 0.026 | 0.75 |
| 0.028 | 0.81 |
| 0.03 | 0.87 |
| 0.032 | 0.93 |
| 0.034 | 0.99 |

**Table 2.** *M* and *n* relation for half normal distribution noise distance

### 3.4 $d_i$ has a uniform distribution

Before sorted, $d_i$ has a uniform distribution. After sorted, the normalized $d_i$ has a beta distribution $B(i, M+1-i)$ [3], which has a distribution function of $C_{M+1}^i x^i (1-x)^{M-i}$.

The normalized $d_{M/2}$ has a distribution function $C_{M+1}^{M/2} x^{M/2} (1-x)^{M/2}$. Based on Hoeffding's inequality [4] with $n = M+1$, $p = 1/2 + M^{a-1}$, and $k = M/2$, we have $I_{1/2-M^{a-1}}(m/2+1, M/2+1) \leq e^{-2M^{2a-1}}$ so

$$P(|d_{M/2} - 1/2| < M^{a-1}) > 1 - e^{-2M^{2a-1}} \tag{42}$$

Similarly, the formulae above should be valid for all $d_i$. Then we have:

$$P\left(\bigcap_{i=1}^M |d_i - i/M| < M^{a-1}\right) > \left(1 - e^{-2M^{2a-1}}\right)^M \tag{43}$$

When $a > 1/2$, the right side goes to 1 when $M$ goes to infinite. Since $2e^{-2M^{2a-1}}$ rapidly reduces to 0, for relatively large $M$, we find an $a$ such that the right side is no less than 99.9%. The minimum $a$-s for the right side is no less than 99.9% for $M$ from 100 to 1000 are listed in the table below:

| M | a |
|---:|---:|
| 100 | 0.696 |
| 200 | 0.676 |
| 300 | 0.666 |
| 400 | 0.660 |
| 500 | 0.655 |
| 600 | 0.652 |
| 700 | 0.649 |
| 800 | 0.647 |
| 900 | 0.645 |
| 1000 | 0.643 |

**Table 3**. *a* for $d_i$ close to the mean *k/M*

When $|d_i - k/M| < M^{a-1}$ for all these distances, for the first group, when $k < 2M^a$, the upper bound is $2M^a$; when $k \geq 2M^a$, the upper bound is:

$$\sum_{i=1}^{k}\left[d_i^p - (d_T - d_i)^p\right]/d_T^p = \sum_{i=1}^{k}\left[\left(\frac{d_i}{d_T}\right)^p - \left(1 - \frac{d_i}{d_T}\right)^p\right]$$

$$\leq \sum_{i=1}^{k-M^a}\left[\left(\frac{(i+M^a)/M}{k/M}\right)^p - \left(1 - \frac{(i+M^a)/M}{k/M}\right)^p\right] + M^a$$

$$\leq \sum_{i=1}^{k-M^a}\left(\frac{i+M^a}{k}\right)^p - \sum_{i=0}^{k-M^a}\left(\frac{i}{k}\right)^p + M^a \quad (44)$$

$$= \sum_{i=k-M^a+1}^{k}\left(\frac{i}{k}\right)^p - \sum_{i=0}^{M^a}\left(\frac{i}{k}\right)^p + M^a$$

$$< 2M^a$$

So the upper bound for TFG is $2M^a$. For TSG, we have:

$$\sum_{i=k+1}^{M}\left[d_i^p - (d_i - d_T)^p\right]/d_T^p = \sum_{i=k+1}^{M}\left[\left(\frac{d_i}{d_T}\right)^p - \left(\frac{d_i}{d_T} - 1\right)^p\right]$$

$$\leq \sum_{i=k+M^a}^{M}\left[\left(\frac{i-M^a}{d_T}\right)^p - \left(\frac{i-M^a}{d_T} - 1\right)^p\right] + M^a$$

(because $x^p - (x-1)^p$ is $decreasing$)

$$\leq \int_{k+1}^{M+1-M^a}\left[\left(\frac{x}{d_T}\right)^p - \left(\frac{x}{d_T} - 1\right)^p\right]dx + M^a$$

$$= \frac{1}{(p+1)d_T^p}\left((M+1-M^a)^{p+1} - (M+1-M^a - d_T)^{p+1} - (k+1)^{p+1}\right) + M^a$$

$$\leq \frac{1}{(p+1)d_T^p}\left((M+1-M^a)^{p+1} - (M+1-M^a - d_T)^{p+1} - d_T^{p+1}\right) + M^a$$

Because

$$-(M+1-M^a - d_T)^{p+1} = -\sum_{i=0}^{\infty}\binom{p+1}{i}(M+1-M^a)^{p+1}\left(-d_T/(M+1-M^a)\right)^i$$

$$= -(M+1-M^a)^{p+1} + (p+1)(M+1-M^a)^p d_T + many \quad negative \quad items$$

, TSG is no larger than:

$$(M+1-M^a)^p d_T^{1-p} - d_T/(1+p) + M^a$$

When $d_T = (1-p^2)^{1/p}M$, it has an upper bound of $p(1-p^2)^{(1-p)/p}(M+1-M^a) + M^a$ Thus, statistically, $L^p (p<1)$ is strict super robust for large *M* and small *p*.

## 3.5 Super Robustness on More General Noise Distribution

Before sorted, $d_i$ has a distribution function $f(x)$ and a cumulative distribution function $F(x)$. Based on order statistics, the distribution of $k$-th smallest value [5] is:

$$C_M^k F(x)^k (1-F(x))^{k-1} f(x) \qquad (45)$$

For this distribution, the probability that $x$ falls in the interval $[a,b]$ is:

$$\int_a^b C_M^k F(x)^k (1-F(x))^{k-1} f(x)dx = \int_{F^{-1}(a)}^{F^{-1}(b)} C_M^k x^k (1-x)^{k-1} dx \qquad (46)$$

Thus, we have:

$$P\left(F^{-1}(k/M - M^a) < d_k < F^{-1}(k/M + M^a)\right) > 1 - 2e^{-2M^{2a-1}}/M \qquad (47)$$

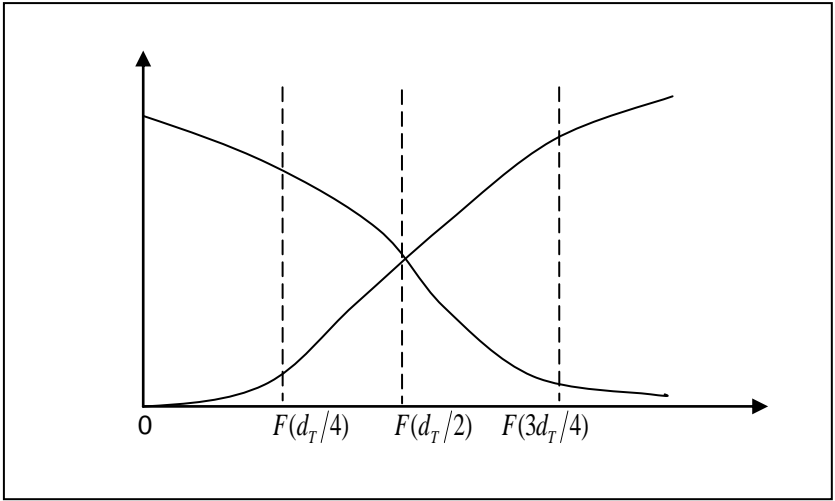

**Fig. 1** The relation of $\left(\dfrac{d_i}{d_T}\right)^p$ and $\left(1-\dfrac{d_i}{d_T}\right)^p$

For TFG,

$$\sum_{i=1}^k \left[d_i^p - (d_T - d_i)^p\right]/d_T^p = \sum_{i=1}^k \left[\left(\dfrac{d_i}{d_T}\right)^p - \left(1-\dfrac{d_i}{d_T}\right)^p\right] \qquad (48)$$

Let's divide it into four groups as shown in Fig. 1: (1) when $d_i \leq F(d_T/4)$,

$$\dfrac{F^{-1}(d_i)}{d_T} \leq 1 - \dfrac{F^{-1}(d_i)}{d_T} - \dfrac{1}{2},$$

or

$$\left(\dfrac{F^{-1}(d_i)}{d_T}\right)^p - \left(1 - \dfrac{F^{-1}(d_i)}{d_T}\right)^p \leq -\dfrac{1}{2^p};$$

when $d_i \in [F(d_T/4), F(d_T/2)]$,

$$\dfrac{F^{-1}(d_i)}{d_T} \leq 1 - \dfrac{F^{-1}(d_i)}{d_T},$$

or

$$\left(\dfrac{F^{-1}(d_i)}{d_T}\right)^p \leq \left(1 - \dfrac{F^{-1}(d_i)}{d_T}\right)^p;$$

when $d_i \in [F(d_T/2), F(3d_T/4)]$,

$$\frac{F^{-1}(d_i)}{d_T} \leq \frac{3}{2} - \frac{F^{-1}(d_i)}{d_T},$$

or

$$\left(\frac{F^{-1}(d_i)}{d_T}\right)^p - \left(1 - \frac{F^{-1}(d_i)}{d_T}\right)^p \leq \frac{1}{2^p};$$

Thus, the overall upper bound of TFG is :

$$(F(d_T) - F(3d_T/4))M - F(d_T/4)M/2^p + (F(3d_T/4) - F(d_T/2))M/2^p. \tag{49}$$

For TSG, when $F(d_T) > 0.5$, the upper bound is $M - MF(d_T)$, in this case, we notice that TFG upper bound is much smaller than $MF(d_T)$; when $F(d_T) \in [0.25, 0.5]$, because the function $x^p - (x-1)^p$ is decreasing, the upper bound is

$$((3/2)^p - (1/2)^p)(1 - F(3d_T/2))M + (F(3d_T/2) - F(d_T))M ;$$ when $F(d_T) \in [0, 0.25]$, the upper bound is $\sum_{i=1}^{2}((i+1)^p - i^p)(F((i+1)d_T) - F(id_T))M + (1 - F(4d_T))M + (F(2d_T) - F(d_T))M$. For small $d_T$, we can obtain a lower upper bound

$$\sum_{i=1}^{\lfloor 1/d_T \rfloor - 1}((i+1)^p - i^p)(F((i+1)d_T) - F(id_T))M + (1 - F(\lfloor 1/d_T \rfloor d_T))M + (F(2d_T) - F(d_T))M$$

by repeating the same estimation approach.

Summarizing the above, statistically, $L^p$ ($p < 1$) is strict super robust for large $M$ and small $p$ when $F(x)$ is "ordinary enough: say, it is a continuous strictly increasing function.

Specially, when all the components of the error vector has a uniform distribution, $d_i$ has a distribution of $f(x) \sim (K+1)x^K$ and $F(x)$ is $x^{K+1}$, for which, statistically, $L^p$ ($p < 1$) is strict super robust for large $M$ and small $p$.

## 4. Strict Super Robustness of Rotation Transform

Let us analyze the super robustness of rotation transform with two-dimensional rotation as an example. Denote that $R(\theta)$ is a rotation. $R(\theta_b)$ is the ideal rotation:

$$|(R(\theta) - R(\theta_b))(I_i)|^p + \sum_{i=1}^{M}|O_{i+n} - R(\theta)(I_{i+n})|^p \geq \sum_{i=1}^{M}|O_{i+n} - R(\theta_b)(I_{i+n})|^p \tag{50}$$

Let $d_i = |O_{i+n} - R(\theta_b)(I_{i+n})|/|I_{i+n}|$. So we expect that $n$ satisfies:

$$n\left|2\sin((\theta - \theta_b)/2)\bar{I}_m\right|^p \geq \sum_{i=1}^{M}\left(|d_i|^p - |(d_i - 2\sin((\theta - \theta_b)/2))|^p\right)|I_{i+n}|^p \tag{51}$$

, where

$$\bar{I}_m = \sum_{i=1}^{n}\min\{I_{ij} \mid j = 1, 2, \ldots, N_{\dim}\}/n$$

Let $d_R = 2\sin((\theta - \theta_b)/2)$. We further make the right side larger to:

$$n\left|d_R \bar{I}_m\right|^p \geq \sum_{i=1}^{M}\left(|d_i|^p - |d_i - d_R|^p\right)|I_{i+n}|^p \tag{52}$$

We also sort $d_i$ so that $d_1 < d_2 < \cdots < d_M$. Similar to above, for a given rotation $R$, divide the right side into two groups $d_i < d_R$ or $d_i > d_R$. For the first group,

$$\sum_{i=1}^{k}\left[d_i^p - (d_R - d_i)^p\right]|I_{i+n}|^p$$

$$\leq \sum_{i=1}^{k}\left[(Si)^p - ((k-i)s)^p\right]$$

$$= (S^p - s^p)\sum_{i=1}^{k-1} i^p + S^p k^p \qquad (53)$$

$$\leq (S^p - s^p)\int_0^k x^p dx + S^p k^p$$

$$= (S^p - s^p)k^{p+1}/(p+1) + S^p k^p$$

$$\leq (S^p - s^p)d_T^{p+1}/s_1^{p+1}(p+1) + S^p d_T^p/s^p$$

, where $S = \max|I_{i+n}|(d_i - d_{i-1})$, $s = \min|I_{i+n}|(d_i - d_{i-1})$, $s_1 = \max(d_i - d_{i-1})$. For the second group, we have that:

$$\sum_{i=k+1}^{M}\left[d_i^p - (d_i - d_R)^p\right]|I_{i+n}|^p \leq \left\{\left[(p+1)d_M^p d_R - d_R^{p+1}\right]/s_1(p+1) + d_R^p\right\}|I_M|^p \qquad (54)$$

So if $n$ satisfies

$$n\overline{|I_m|^p}$$

$$\geq (S^p - s^p)d_R/s_1(p+1) + (S/s)^p + \left|(p+1)d_M^p d_R - d_R^{p+1}\right||I_M|^p / \left[d_R^M s_1(p+1)\right] + |I_M|^p$$

$$= r^p + p\left[d_M^p d_R^{1-p}|I_M|^p + \left((r^p-1)|\bar{I}_m|^p - |I_M|^p\right)d_R/(p+1)\right]/s_1 + |I_M|^p$$

, we have $D_R > D_{R_b}$.

When $d_R = \left[(1-p^2)/\left(|I_M|^p + (1-r^p)|\bar{I}_m|^p\right)\right]^{1-p} |I_M| d_M$

$n$ reaches the minimum:

$$n > \left\{p\left[(1-p^2)/|I_M|^p + (1-r^p)|\bar{I}_m|^p\right]^{(1-p)/p} |I_M| d_M \Big/ s_1 + |I_M|^p + r^p\right\}\Big/\overline{|I_m|^p} \qquad (55)$$

For the rotation in a higher dimension, it can be decomposed into $N_{\dim}(N_{\dim}-1)/2$ two-dimensional rotations. If for each rotation, the error distances are all different, we have a minimum $n_i$ for each $i = 1, 2, \cdots, N_{\dim}(N_{\dim}-1)/2$, so the lower bound of $n$ to guarantee that the ideal transform reaches the minimum is

$$n = \max\{n_i \mid i = 1, 2, \cdots, N_{\dim}(N_{\dim}-1)/2\} \qquad (56)$$

Another approach to understand the robustness is that because $|(R(\theta_1) - R(\theta_2))I_i| < 2$, so when all the error distances are larger than 2,

$$\sum_{i=k+1}^{M}\left[d_i^p - (d_i - d_R)^p\right]|I_{i+n}|^p \leq |2I_M|^p M \qquad (57)$$

For a given set of input, we find an upper bound of $p$ such that $|2I_M|^p < (1+M)/M$. This means when we always have a $p$ which allows us to obtain the ideal rotation.

## 5. Robustness on Euclidean Transform

When $T$ is an Euclidean transform, that is, a rotation (rotation first) and a translation, with the condition that $|I_i|$ has an upper limit, we have that when the translation item of $T$ is large enough relative to $\sum |I_i|$, the difference introduced by the rotation item is relatively small. Mathematically, we have $\left\| T(I_i) - T_b(I_i) \right| / \left| a_T - a_{T_b} \right| - 1 \right| < \varepsilon$; when $\left| a_T - a_{T_b} \right| \to \infty, \ \varepsilon \to 0 \ \left| a_T - a_{T_b} \right| \to \infty, \ \varepsilon \to 0$.

Thus $n > \sum_{i=n+1}^{M+n} \left( \left| O_i - T_b(I_i) \right|^p - \left| O_i - T(I_i) \right|^p \right) / \left[ (1-\varepsilon) \left| a_T - a_{T_b} \right|^p \right]$

Or we expect that

$$n > \sum_{i=n+1}^{M+n} \left( \left| (T_b - T)(I_i) \right|^p \right) / \left[ (1-\varepsilon) \left| a_T - a_{T_b} \right|^p \right]$$
or
$$n > \sum_{i=1}^{M} \left( (1+\varepsilon) \left| a_T - a_{T_b} \right|^p \right) / \left( (1-\varepsilon) \left| a_T - a_{T_b} \right|^p \right) = (1+\varepsilon) M / (1-\varepsilon)$$
(58)

This exactly means $L^p$ is a robust estimator on Euclidean transformation when the noises are very large.

## 6. Super Robustness on Euclidean Transform

Similar to above, we obtain a similar super robustness analysis of $L^p$ for Euclidean transformation: For the number of small amount of ideal observation will generate an estimation that doesn't move to infinite when the noise observations go to infinite:

$$n > \left\{ p \left[ (1-p^2)/(2-R^p) \right]^{(1-p)/p} d_M / s + 1 + R^p \right\} (1+\varepsilon)/(1-\varepsilon) \tag{59}$$

## 7. Strict Robustness on Scaling Transform

For a scaling transform $T(I) = s_T I$, where the scaling factor $s_T$ is a positive real number. That $D_T > D_{T_b}$ means:

$$\sum_{i=1}^{n} \left| O_i - T(I_i) \right|^p + \sum_{i=1}^{M} \left| O_{i+n} - T(I_{i+n}) \right|^p \geq \sum_{i=1}^{M} \left| O_{i+n} - T_b(I_{i+n}) \right|^p$$

Thus, $\sum_{i=1}^{n} \left| O_i - T(I_i) \right|^p + \sum_{i=1}^{M} \left| O_{i+n} - T(I_{i+n}) \right|^p = \sum_{i=1}^{n} \left| (s_{T_b} - s_T) I_i \right|^p + \sum_{i=1}^{M} \left| O_{i+n} - s_T I_{i+n} \right|^p$ Since

Because $(a+b)^p < a^p + b^p$, we have:

$$\sum_{i=1}^{n} \left| T(I_i) - T_b(I_i) \right|^p > \sum_{i=1}^{M} \left| O_{i+n} - T_b(I_{i+n}) \right|^p - \sum_{i=1}^{M} \left| O_{i+n} - T(I_{i+n}) \right|^p \tag{60}$$

If $\sum_{i=1}^{n} \left| (s_{T_b} - s_T) I_i \right|^p > \sum_{i=1}^{M} \left| (s_{T_b} - s_T) I_{i+n} \right|^p$ or $\sum_{i=1}^{n} |I_i|^p > \sum_{i=1}^{M} |I_{i+n}|^p$

then
$$D_T > D_{T_b}$$

This means that when the ideal observation group controls the noise group, $L^p$ obtains the ideal estimation, no matter the relation of the size of the ideal observation group and the noise observation group. This interesting result is different with the other results derived above. This is not a very useful result since we don't know the relation of the good observation and noisy observation in almost all the cases.

The right side of (64) can be enlarged to:

$$\sum_{i=1}^{M}\left(\||O_{i+n}|+|T_b(I_{i+n})|\|^p - \||O_{i+n}|-|T(I_{i+n})|\|^p\right)$$

$$\leq \sum_{i=1}^{M} p\||O_{i+n}|+|T_b(I_{i+n})|\|^{p-1}(|T_b(I_{i+n})|+|T(I_{i+n})|)$$

$$= \sum_{i=1}^{M} p\||O_{i+n}|+|T_b(I_{i+n})|\|^{p-1}(|s_T|+|s_{T_b}|)$$

So when $\sum_{i=1}^{n}|I_i|^p \geq \sum_{i=1}^{M} p\||O_{i+n}|+|T_b(I_{i+n})|\|^{p-1}(|s_T|+|s_{T_b}|) / (|s_T|-|s_{T_b}|)$, we have $D_T > D_{T_b}$

Without lose generality, let $s_{T_b}=1$, when $s_T > 2$ and $n \geq \sum_{i=1}^{M} 2p\||O_{i+n}|+|I_{i+n}|\|^{p-1}/|\bar{I}_n|$, we have $D_T > D_{T_b}$. When $O_{i+n} \to \infty$, $\sum_{i=1}^{M} 2p\||O_{i+n}|+|I_{i+n}|\|^{p-1}/|\bar{I}_n| \to 0$. Thus, $L^p$ has super robustness on scaling transformation.

Another way to investigate the robustness of $L^p$ on the scaling transform: the scaling transform on $I_i$ is a translation on $\log(I_i)$; thus all the results on translation are valid on scaling transform when we estimate the translation of $\log(I_i)$ using maximum likelihood estimation of $L^p$.

## 8. Experiments

This section shows two experiments using the approach described above. A simplex programming approach is used to find the maximum likelihood estimation.

### 8.1 2D Pattern Match

The first experiment is matching two 2D pattern shown in Figure 1 (a) and (b):

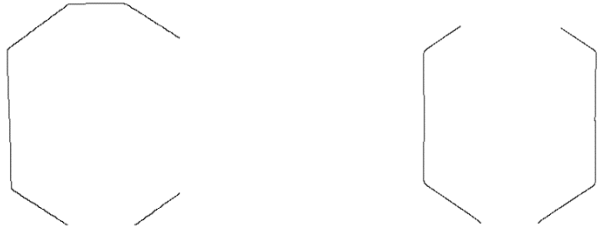

(a)          (b)
**Figure 1**. The original patterns to be matched

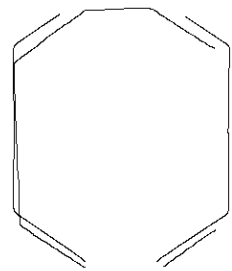
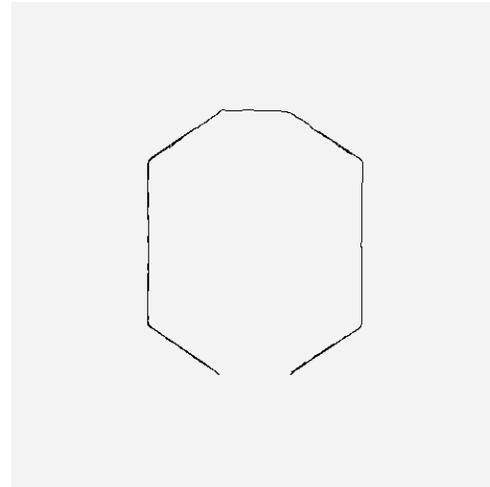

(a) *p*=2.0  (b) *p*=0.8

**Figure 2**. The matched results with p=2.0 (left) and p= 0.8 (right)

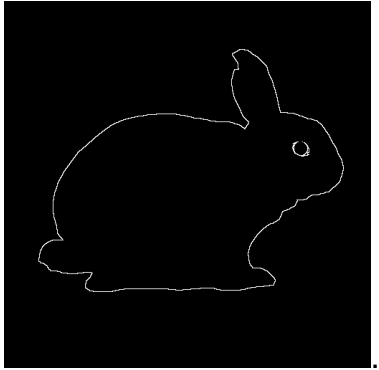
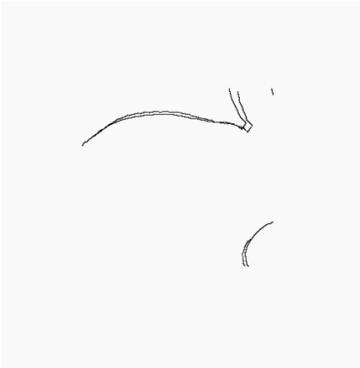
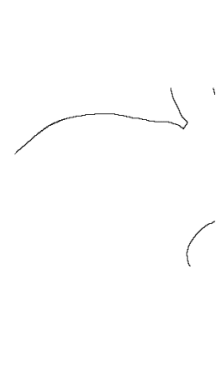

(a) Original image  (b) p=2.0  (c)p=0.8

**Figure 3**. 2D pattern matching

### 3.1 2D Point Set Match

We did experiments on 2D point set matching: in the experiments, random 2D point sets are generated and the majority of the points are moved with a random 2D translation; the point set with noisy are transformed with a Euclidean transform; the transformed noise point set is matched with the original point sets. The below is an example: (a) The 30 blue dots are the original points; the 25 red points are the transformed points with noise, and the 5 green points are the transformed points with no noise; (b) The match results with *p*=2.0; (c) The match results with *p*=1.0; and (d) The match results with *p*=0.5. It is easy to see that with *p*=0.5, the ideal transformation is estimated.

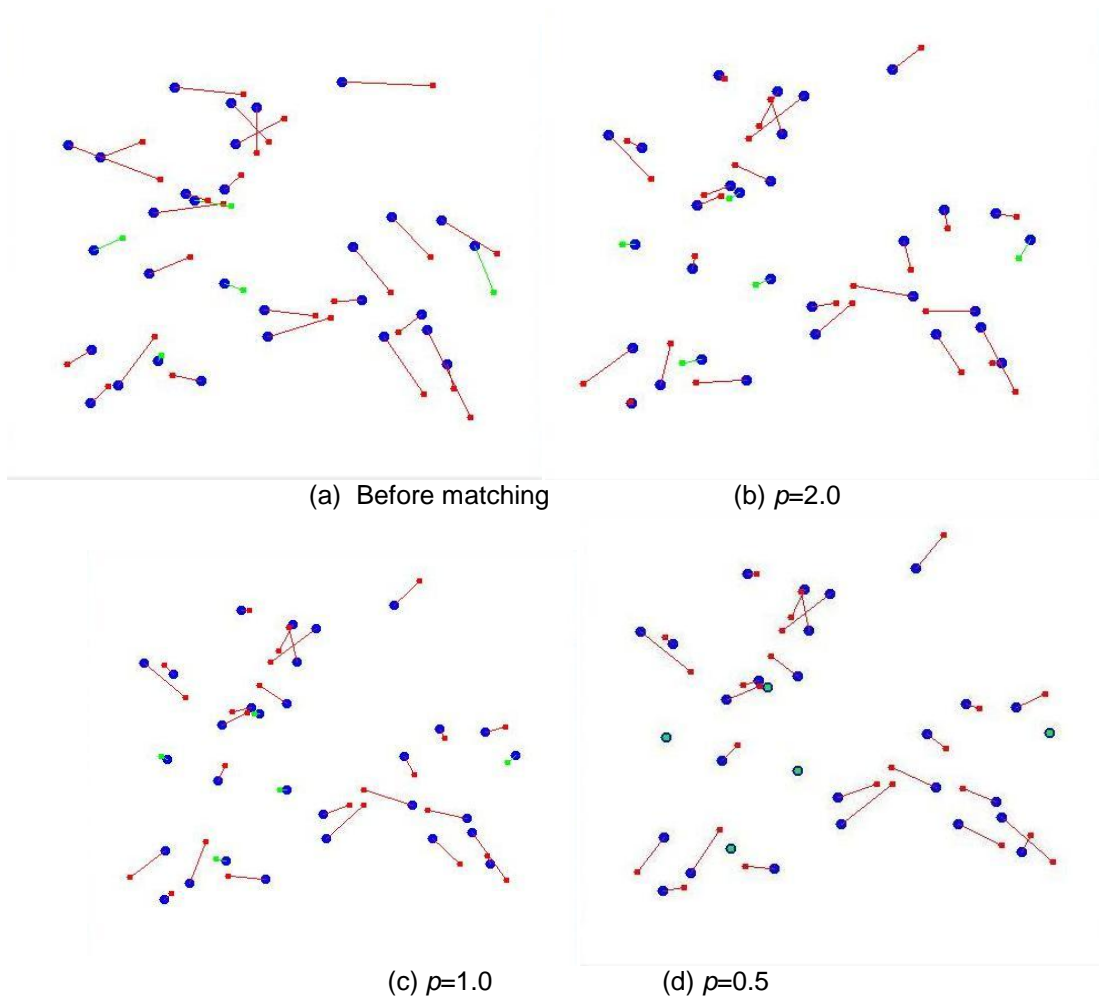

(a) Before matching  (b) $p=2.0$

(c) $p=1.0$  (d) $p=0.5$

**Fig. 4** 2D Point Set Matching

## 9. Conclusions

The [strict] super robustness and robustness of the maximum likelihood estimator of the exponential power distribution, or $L^p$ estimation, where $p<1$, are theoretically proofed. Also, super robustness are observed in experiments of pattern matching. The proposed approach will be not only useful for pattern matching but for any estimation problems with very noisy data.

## 10. References

[1]. *Robust Statistics*, Peter. J. Huber, Wiley, 1981 (republished in paperback, 2004)

[2]. *Robust Statistics - The Approach Based on Influence Functions*, Frank R. Hampel, Elvezio M. Ronchetti, Peter J. Rousseeuw and Werner A. Stahel, Wiley, 1986 (republished in paperback, 2005).

[3]. http://en.wikipedia.org/wiki/Beta_distribution

[4].http://en.wikipedia.org/wiki/Binomial_distribution#Cumulative_distribution_function

[5]. http://en.wikipedia.org/wiki/Order_statistic